\documentclass[conference]{IEEEtranSPMB}

\IEEEoverridecommandlockouts
\ifCLASSINFOpdf
\else
\fi
\usepackage{amsmath,graphicx,soul,subcaption,balance}
\usepackage[numbers,sort&compress]{natbib}

\usepackage{parskip}
\usepackage{float} 
\usepackage[all]{nowidow}
\usepackage{mathptmx}

\usepackage{multirow} 

\usepackage{color}
\usepackage{soul,mathrsfs}
\usepackage[utf8]{inputenc}
\usepackage[T1]{fontenc}
\usepackage{array}
\usepackage[hyphens, spaces, obeyspaces]{url}

\usepackage[papersize={8.5in,11in}, left=1in, right=1in, top=1in, bottom=1in]{geometry}
\newcolumntype{L}[1]{>{\raggedright\let\newline\\\arraybackslash\hspace{0pt}}m{#1}}
\newcolumntype{C}[1]{>{\centering\let\newline\\\arraybackslash\hspace{0pt}}m{#1}}
\newcolumntype{R}[1]{>{\raggedleft\let\newline\\\arraybackslash\hspace{0pt}}m{#1}}

\usepackage[singlelinecheck=false,justification=centering]{caption}
\DeclareCaptionLabelFormat{mylabel}{#1 #2.\hspace{0.7ex}}
\renewcommand{\thetable}{\arabic{table}}

\usepackage{xcolor,colortbl}
\definecolor{light-gray}{gray}{0.83}

\newcommand{\spmbtitlefont}{\fontsize{11.0pt}{11.00pt}\selectfont\bf\vspace{0.7em}}
\newcommand{\spmbauthorfont}{\fontsize{11.0pt}{11.0pt}\selectfont\vspace{0em}}
\newcommand{\subparagraph}{}
\usepackage[compact]{titlesec}
\titleformat{\section}
   {\center\normalfont\sc}{\thesection.}{0.7ex}{}
\titlespacing{\section}{0pt}{2ex}{1.5ex}
\titlespacing{\subsection}{0pt}{1.5ex}{1.2ex}
\titlespacing{\subsubsection}{0pt}{1ex}{0.9ex}
\makeatletter
\renewcommand*{\@seccntformat}[1]{\csname the#1\endcsname .\hspace{0.7em}}
\makeatother

\usepackage{fancyhdr,lastpage}
\fancypagestyle{firststyle}
{
   \fancyhf{}
   \fancyfoot[L]{\scriptsize 979-8-3315-7370-6/25/\$31.00 \copyright2025 IEEE}
   \fancyfoot[R]{\scriptsize December 6, 2025}
   \fancyfoot[C]{\scriptsize IEEE SPMB 2025}
}
\title{\spmbtitlefont ECG Latent Feature Extraction with Autoencoders for Downstream Prediction Tasks
{\vspace{-2.4\baselineskip}
}
}
    \author{\spmbauthorfont\IEEEauthorblockN{
    Christopher Harvey\textsuperscript{\it 1}, 
    Sumaiya Shomaji\textsuperscript{\it 2}, 
    Zijun Yao\textsuperscript{\it 2}, 
    Amit Noheria\textsuperscript{\it 1}, 
    }
    \vspace{0.9em}
    \IEEEauthorblockA{\spmbauthorfont 
        1. Department of Cardiovascular Medicine, The University of Kansas Medical Center,\\ Kansas City, Kansas 66160 USA \\
        2. Department of Electrical Engineering and Computer Science, The University of Kansas,\\ Lawrence, Kansas 66045 USA \\
        cjh@ku.edu, shomaji@ku.edu, zyao@ku.edu, noheriaa@gmail.com
    }
}

\newcommand{\PaperTitleSummary}{C.\ Harvey, et al.: Compairson of Autoencoder ...}

\begin{document}

\IEEEaftertitletext{}
\maketitle

\begin{abstract}

The electrocardiogram (ECG) is an inexpensive and widely available tool for cardiac assessment.
Despite its standardized format and small file size, the high complexity and inter-individual variability of ECG signals (typically a 60,000-size vector with 12 leads at 500 Hz) make it challenging to use in deep learning models, especially when only small training datasets are available.
This study addresses these challenges by exploring feature generation methods from representative beat ECGs, focusing on Principal Component Analysis (PCA) and Autoencoders to reduce data complexity.
We introduce three novel Variational Autoencoder (VAE) variants—Stochastic Autoencoder (SAE), Annealed $\beta$-VAE (A$\beta$-VAE), and Cyclical $\beta$-VAE (C$\beta$-VAE)—and compare their effectiveness in maintaining signal fidelity and enhancing downstream prediction tasks using a Light Gradient Boost Machine (LGBM).
The A$\beta$-VAE achieved superior signal reconstruction, reducing the mean absolute error (MAE) to 15.7$\pm$3.2 $\mu$V, which is at the level of signal noise.
Moreover, the SAE encodings, when combined with traditional ECG summary features, improved the prediction of reduced Left Ventricular Ejection Fraction (LVEF), achieving an holdout test set area under the receiver operating characteristic curve (AUROC) of 0.901 with a LGBM classifier.
This performance nearly matches the 0.909 AUROC of state-of-the-art CNN model but requires significantly less computational resources. Further, the ECG feature extraction-LGBM pipeline avoids overfitting and retains predictive performance when trained with less data.
Our findings demonstrate that these VAE encodings are not only effective in simplifying ECG data but also provide a practical solution for applying deep learning in contexts with limited-scale labeled training data.
\end{abstract}

\begin{keywords}
Electrocardiogram, Dimensionality Reduction, Variational Autoencoder, Signal Processing
\end{keywords}

\IEEEpeerreviewmaketitle    
\thispagestyle{firststyle}  
\section{Introduction}
\label{sec:intro}

The electrocardiogram (ECG) is a non-invasive clinical tool that records the heart's electrical activity via electrodes on the skin, typically as a 12-lead ECG over 10 seconds at 500 Hz, producing 60,000 data points. The 10-sec ECG spans 8-17 cardiac cycles at usual heart rates. Each cycle includes P wave (atrial depolarization), QRS complex (ventricular depolarization), and T wave (ventricular repolarization), with characteristics like amplitude and duration varying heavily across leads and different individuals, reflecting differences in cardiac structure and function.
The P wave is short, low frequency and low amplitude; QRS complex is short, high frequency and high amplitude while the T wave is long, low frequency and intermediate amplitude.
These waves and intervals between cardiac cycles are interspersed with periods of zero electrical activity (or electrical baseline) which makes the signal data distribution very skewed.
All humans have unique hearts with variations in size, anatomy and electrophysiology, accounting for the interindividual differences in the ECG signal.
E.g., one person might have a QRS complex amplitude of 0.25 mV and another 5 mV.
The morphology of each person's ECG can be very different, e.g., an ECG lead of one person might have a smooth monophasic positive QRS complex (R wave) while another might have a notched R wave while another has triphasic Q-R-S deflections.
Age, sex, body structure, the specific lead in question and cardiac diseases all affect the morphology of the ECG.
The complexity of ECG signals poses challenges for Deep Learning (DL) due to high variance in wave morphology, skewed data distribution, and temporal volatility. 
This complexity and diversity makes it challenging for DL models to generalize from and requires large training datasets to avoid overfitting. This problem of inadequate training ECG samples for DL is especially encountered for rare clinical outcomes like myocardial infraction or invasive procedures like catheter ablation.

Traditional ECG simplification using summary statistics (e.g., heart rate, QRS duration, etc. \cite{berkaya2018survey}) fails to capture nuanced morphology differences. For instance, two ECGs could have identical summary statistics but exhibit completely different morphologies, such as Left Bundle Branch Block (LBBB) versus Right Bundle Branch Block (RBBB). Prior studies, like Kumar and Chakrapani's PCA-based approach \cite{kumar2022classification} or Dasan and Panneerselvam's convolutional denoising Autoencoder (AE) with LSTM \cite{DASAN2021102225}, reduce ECG dimensionality but often miss non-linear relationships critical for diagnosis.
Variational Autoencoders (VAEs) have been used for data augmentation, but can also address these limitations by capturing non-linear features and providing a structured latent space for better generalization. We propose a VAE-based framework with three novel variants—Stochastic Autoencoder (SAE), Cyclical $\beta$-VAE (C$\beta$-VAE), and Annealed $\beta$VAE (A$\beta$-VAE)—to optimize ECG latent representations for high-fidelity signal reconstruction and improved predictive performance, esp. with limited-size training datasets. These variants balance reconstruction fidelity and latent space regularization, enabling DL on limited ECG datasets and facilitate integration with simpler algorithms like tree-based models, offering an alternative to standard DL methods like Convolutional Neural Networks (CNNs).

\section{Methods}

\subsection{Data}
\begin{figure}[ht]  
  \centering
  \includegraphics[width=1.\linewidth \vspace{-5pt}]{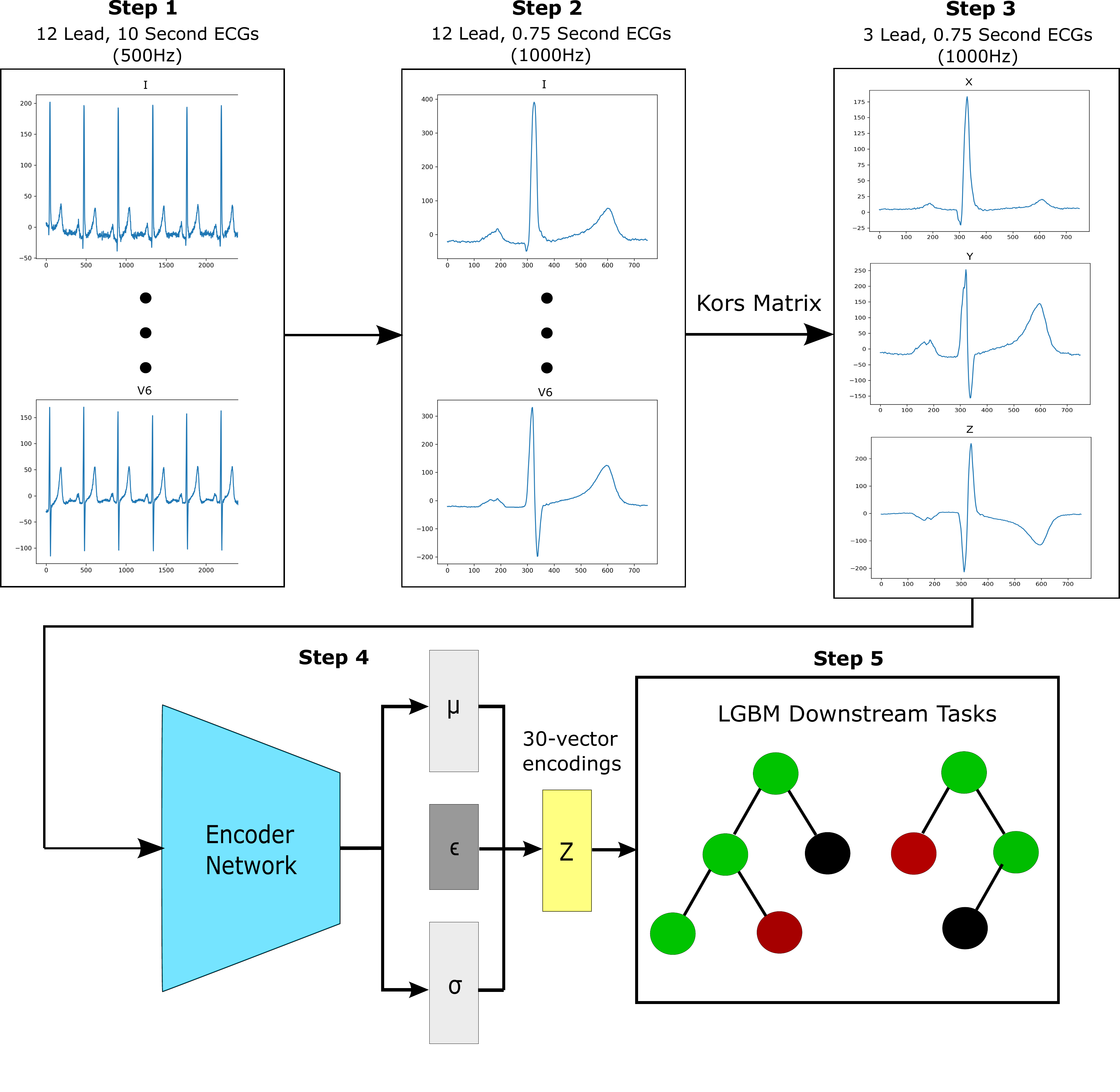}
  
  \caption{Data pipeline for downstream prediction tasks. Converting from 10 s data to 0.75 s to X,Y,Z beats to 30-vector encoding.}
  \label{fig:pipeline}
\end{figure}

We reduced 10-sec ECG recordings to a 750 ms representative beat, centered 100 ms after the QRS complex onset, capturing key morphological features while shrinking data size by eliminating the redundant multiple cardiac cycles. Then, using Kors’s conversion matrix \cite{kors1990reconstruction}, we transformed the eight independent ECG leads (I, II, V1-V6) into three orthogonal X (right to left), Y (cranial to caudal), and Z (anterior to posterior) leads, reducing the 1000 Hz 120,000-datapoint 10-sec 12-lead ECG to a 2250-datapoint 750 ms 3-lead ECG as shown in Figure \ref{fig:pipeline}.

For training stability, we applied global absolute max scaling, normalizing signals to [-1, 1] by dividing by the dataset’s maximum absolute amplitude, followed by mean subtraction to remove baseline offsets. 
Unlike z-score normalization (per-patient or global), which reduces inter-patient amplitude differences, absolute max scaling preserves these clinically important relative differences. This approach combines the centering benefits of z-score normalization with the preservation of intra- and inter-patient amplitude ratios critical for ECG analysis. 
This project was approved by The University of Kansas Medical Center IRB (STUDY00160252). Our dataset comprised of 1,065,368 twelve-lead ECGs obtained between 2008-2022 was split by unique patient IDs into training (90\%) and test (10\%) sets. ECGs were linked to echocardiographic data when available on querying Healthcare Enterprise Repository for Ontological Narration (HERON) \cite{Murphy2010}\cite{Waitman2011}.

\subsection{Overview of the Models}

We compare 7 models to reduce the dimensionality of ECG data: PCA, AE, SAE, VAE, $\beta$-VAE, C$\beta$-VAE, and A$\beta$-VAE. The SAE, C$\beta$-VAE, and A$\beta$-VAE are novel implementations of VAEs, each with unique features tailored to address challenges in ECG data encoding. 

The PCA model was trained using the Incremental PCA method from the sci-kit learn library. It was configured to produce 30 components to match the 30 latent encodings generated by the VAEs, enabling a fair comparison between different methods. The AE and the VAE variants (SAE, VAE, $\beta$VAE, C$\beta$-VAE, A$\beta$-VAE) were trained using the same architecture, with a tailored loss function specific to each model, as discussed in subsequent sections.

\subsection{Overview of VAEs}

\begin{figure}  
  \centering
  \includegraphics[width=0.8\linewidth]{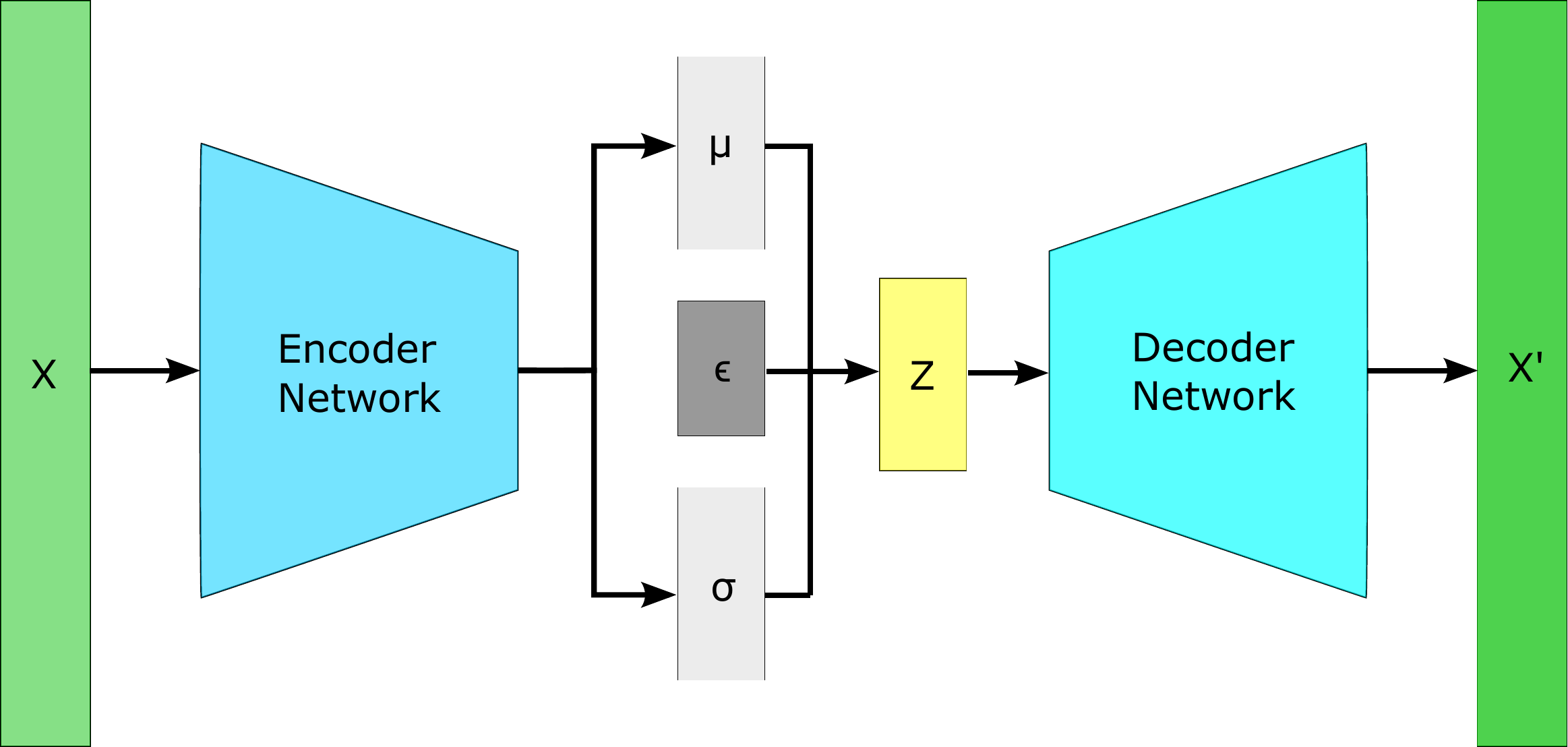}
  
  \caption{Overview of VAE architecture}
  \label{fig:modelarch}
\end{figure}

All VAE models featured an encoder-decoder structure built with Convolutional Neural Networks (CNNs). The encoder included four 2D convolutional layers (filters: 256, 256, 512, 512) with a filter width of 9, stride of 2, TanH activations, and batch normalization, followed by two fully connected layers with L2 regularization (0.01) and dropout (0.25). The latent variable z was sampled from a Gaussian distribution (except the SAE, which is a stochastic distribution) via the reparameterization trick:
\begin{equation}
z = \mu + \sigma \odot \epsilon
\end{equation}
Where z is the latent variable encoding, $\mu$ is the population mean, $\sigma$ is the standard deviation of the population distribution, and $\epsilon$ is a small randomly sampled value to allow for back propagation via the reparameterization trick \cite{kingma2013auto}.
The decoder mirrored the encoder with two fully connected layers and four transpose convolutional layers (filters: 512, 256, 128, 3) to reconstruct the signal. Models were trained on \url{~}1.1 million ECGs using TensorFlow, Adam optimizer (learning rate: 0.000001), 50 epochs, and a batch size of 32, on a GeForce RTX 3090 with 128 GB RAM and an AMD Ryzen 9 3900XT CPU, though inference is lightweight and hardware-agnostic. The general model architecture can be seen in Figure \ref{fig:modelarch}.

\subsection{Modified ELBO Loss}

The reconstruction process is guided by an Evidence Lower Bound (ELBO) loss function that combines a weighted Mean Squared Error (MSE) between the original and reconstructed signals, a Kullback–Leibler (KL) divergence term to regularize the latent space and promote feature disentanglement \cite{kullback1951information}. A smooth latent space is useful if the VAE is used for augmented data generation. A $\beta$ term is used to control the balance between reconstruction quality and latent space regularization. Each 250-ms section of the signal (P wave, QRS complex, and T wave) has unique weights that are selected to diminish the effects of differences in their amplitudes. This allows the model to focus evenly across all the waves and accurately reconstruct all of them without giving priority to the higher amplitude QRS complex. 

The loss function $L_\beta$ is given below
\begin{equation}
L_\beta = L_E + \beta KL
\end{equation}
\begin{equation}
L_E = \theta_{\text{P}} \cdot L_{\text{P}} + \theta_{\text{QRS}} \cdot L_{\text{QRS}} + \theta_{\text{T}} \cdot L_{\text{T}}
\end{equation}
\begin{equation}
KL = -0.5 \left(1 + \ln(\sigma^2_z) - (\mu_z)^2 - e^{\ln(\sigma^2_z)} \right)
\end{equation}
Where $L_\beta$ is the total loss of the model by which the model's gradient is updated.
Where $L_E$ is a weighted MSE for the P, QRS, and T wave segments where $L_{\text{P/QRS/T}}$ is the MSE between the input $x$ and its reconstruction $x'$ for each segment, and $KL$ is the KL divergence for latent space regularization. We used weights, $\theta_{\text{P}}=20.0$, $\theta_{\text{QRS}}=10.0$ and $\theta_{\text{T}}=15.0$, to balance reconstruction across the segments, countering the QRS complex’s higher amplitude, which would otherwise dominate training loss using unweighted MSE. The KL term ensured smoothness in the encodings and promote feature disentanglement. $\beta$ = 3 was selected for $\beta$-VAE to enhance a smooth latent space structure.

\subsection{Overview of Novel Variants}

For the C$\beta$-VAE, we implemented a cyclical annealing schedule where the values of $\beta$ range from 0 to 5, changing each epoch during training.
This cycling means that the model alternates between three different loss functions: AE ($\beta$ = 0), VAE ($\beta$ = 1), and $\beta$-VAE ($\beta$\textgreater1).
At $\beta$=0, the model is essentially just an AE where the loss function only includes the reconstruction error, $L_E$.
At $\beta$=1, the model is a regular VAE with a KL loss and error term\cite{kingma2013auto}.
At $\beta$\textgreater1, there is an additional term added to increase the model’s focus on disentanglement by enhancing the effect of KL on the total loss\cite{higgins2017betavae}.
By cycling from 0 to 5 we make the model go through periods of focusing purely on reconstruction and periods where the model focuses more on understanding the abstract interaction between the data points.
The original paper which introduced cyclical annealing KL loss \cite{fu2019cyclical} had the $\beta$ term cycle between 0 and 1.
They also had $\beta$ hard reset back to 0 from 1.
Instead, we propose to have the $\beta$ term go between 0 and 5 without a hard reset.
$\beta$ goes from 0 to 5 in 10 epochs and 5 to 0 in 10 epochs with a complete cycle  every 20 epochs.
This cyclical annealing approach accelerates convergence, enabling the model to produce usable reconstructions within just 10 epochs.

The A$\beta$-VAE, on the other hand, utilizes a reverse annealing process.
In this variant, the $\beta$ value starts at 10 and is gradually reduced to 0 over the course of 50 epochs.
This process allows the model to initially focus on disentanglement and the structure of the latent space before shifting its emphasis toward reconstruction.
By beginning with a high $\beta$ value, the A$\beta$-VAE emphasizes the regularization of the latent space, which can lead to more meaningful and distinct features in the encoded representations.
This produced a model which was exceptionally good at reconstruction fidelity.

For the SAE, the value of $\beta$ was set to 0, which omits the KL term altogether.
This essentially creates an AE with a stochastic latent distribution, z, which is only trained using the reconstruction loss.
So, instead of learning a Gaussian distribution from the $KL$ loss, it learns the distribution which minimizes the reconstruction loss.
The purpose of the VAE architecture was to create a Gaussian distribution for data generation purposes.
The SAE is a counter-intuitive departure from that to create an encoder which only focuses on the reconstruction of the ECG signal.
The SAE allows for a more flexible and data-driven approach to encoding which improves the performance of downstream tasks while retaining some benefits of VAEs.

\begin{table*} [h]
\caption{MAE, MSE, and DTW for different models for representative beat X, Y, Z-lead ECG reconstructions (N=1,065,368)}
\centering
\scriptsize
\begin{tabular}{lcccccc}
\hline
Model & 1st 250 ms (p wave) & 2nd 250 ms (QRS) & 3rd 250 ms (T wave) & Full signal MAE & Full signal MSE & Full signal DTW \\
& MAE ($\mu V$) Avg $\pm$ SD & MAE ($\mu V$) Avg $\pm$ SD & MAE ($\mu V$) Avg $\pm$ SD & MAE ($\mu V$) Avg $\pm$ SD & ($\mu V^2$) Avg $\pm$ SD & Avg $\pm$ SD \\
\hline
PCA & 19.1$\pm$6.5 & 29.5$\pm$7.3 & 22.7$\pm$7.7 & 24.0$\pm$5.0 & 1842.9$\pm$840.0 & 667.9$\pm$218.7 \\
AE & \textbf{11.2$\pm$3.0} & 23.2$\pm$5.5 & \textbf{12.8$\pm$3.7} & 15.8$\pm$3.1 & 739.8$\pm$316.1 & 313.7$\pm$94.0 \\
SAE & 11.4$\pm$2.7 & 31.7$\pm$8.1 & 12.7$\pm$3.4 & 18.7$\pm$3.7 & 1131.3$\pm$514.8 & 387.3$\pm$131.8 \\
VAE & 11.9$\pm$2.7 & 28.9$\pm$6.9 & 12.8$\pm$3.3 & 17.9$\pm$3.5 & 996.2$\pm$433.2 & 361.2$\pm$115.5 \\
$\beta$-VAE & 11.5$\pm$3.0 & 23.6$\pm$5.6 & 13.1$\pm$3.7 & 16.2$\pm$3.2 & 755.6$\pm$325.6 & 317.5$\pm$94.8 \\
A$\beta$-VAE & \textbf{11.2$\pm$3.0} & \textbf{22.6$\pm$5.3} & \textbf{12.8$\pm$3.8} & \textbf{15.7$\pm$3.2} & \textbf{701.6$\pm$304.8} & \textbf{308.1$\pm$92.1} \\
c$\beta$-VAE & 12.0$\pm$2.6 & 31.9$\pm$6.6 & 14.0$\pm$3.8 & 19.3$\pm$3.4 & 1202.7$\pm$491.3 & 400.6$\pm$130.3 \\
\hline
\end{tabular}
\label{tab:MAE_MSE_DTW}
\end{table*}

\begin{table*}[h] 
\caption{Comparison of MAE and DTW for X, Y, and Z signals across models (N=97,464)}
\centering
\begin{tabular}{lcccccc}
\hline
Model & X Signal MAE ($\mu V$) & Y Signal MAE ($\mu V$) & Z Signal MAE ($\mu V$) & X Signal DTW & Y Signal DTW & Z Signal DTW \\
\hline
PCA & 25.8$\pm$21.4 & 25.6$\pm$20.7 & 23.5$\pm$21.0 & 751.9$\pm$856.4 & 715.8$\pm$778.0 & 647.2$\pm$766.0 \\
AE & 16.3$\pm$13.4 & 16.9$\pm$13.1 & 16.5$\pm$13.2 & 338.0$\pm$412.4 & 337.0$\pm$384.9 & 332.6$\pm$390.8 \\
SAE & 19.7$\pm$12.8 & 21.1$\pm$12.7 & 17.1$\pm$11.9 & 429.5$\pm$404.8 & 455.9$\pm$406.5 & 340.0$\pm$362.3 \\
VAE & 18.9$\pm$14.0 & 20.5$\pm$14.4 & 16.4$\pm$11.8 & 400.4$\pm$413.1 & 419.5$\pm$400.1 & 327.9$\pm$349.7 \\
$\beta$-VAE & 16.9$\pm$14.0 & 17.8$\pm$14.0 & \textbf{15.9$\pm$12.6} & 346.5$\pm$413.8 & 350.4$\pm$390.2 & 321.4$\pm$376.6 \\
A$\beta$-VAE & \textbf{16.1$\pm$13.5} & \textbf{16.9$\pm$12.9} & 16.0$\pm$13.1 & \textbf{334.3$\pm$411.2} & \textbf{335.8$\pm$384.4} & \textbf{321.7$\pm$381.1} \\
c$\beta$-VAE & 20.6$\pm$14.9 & 22.0$\pm$15.2 & 17.8$\pm$11.5 & 438.2$\pm$425.6 & 462.3$\pm$422.5 & 365.5$\pm$358.2 \\
\hline
\end{tabular}
\label{tab:MAE_DTW_Signals}
\end{table*}

\subsection{Comparison of Models for Downstream Prediction}

To compare downstream clinically relevant predictions, we used prediction of reduced Left Ventricular Ejection Fraction (LVEF $\leq$35\%). LVEF is the percent of blood that is pumped by the left ventricle with each contraction and reduced LVEF is the defining feature of Heart Failure with Reduced Ejection Fraction. 
The baseline comparator CNN trained on the full 10-sec 12-lead ECG data we used for this paper is our implementation of the CNN model architecture from Mayo Clinic\cite{attia2019screening}.
This is a state-of-the-art CNN architecture which has been used to model a variety of targets (sex, age, LVEF, etc\cite{attia2019screening}\cite{attia2019age}).
The performance of our CNN model for reduced LVEF in our test set has AUROC 0.909 which is in the range published by other centers \cite{cho2021artificial}.

We compare the performance of our baseline CNN model trained on raw ECG signal data with Light Gradient Boosted Machine (LGBM) models \cite{10.5555/3294996.3295074} trained on the encodings from our AE models.
For the LGBM models, we only altered 8 parameters from their default values: Max Depth (15), Colsample Bytree(.9), Extra Trees (True), Top K (100), Learning Rate (0.1), Num Estimators (1,000,000 with early stopping), Reg Alpha/Lambda (0.95).

\section{Experiments and Results}
\subsection{Signal Reconstruction}

\begin{figure*}[h]  
  \centering
  \includegraphics[width=1.\linewidth]{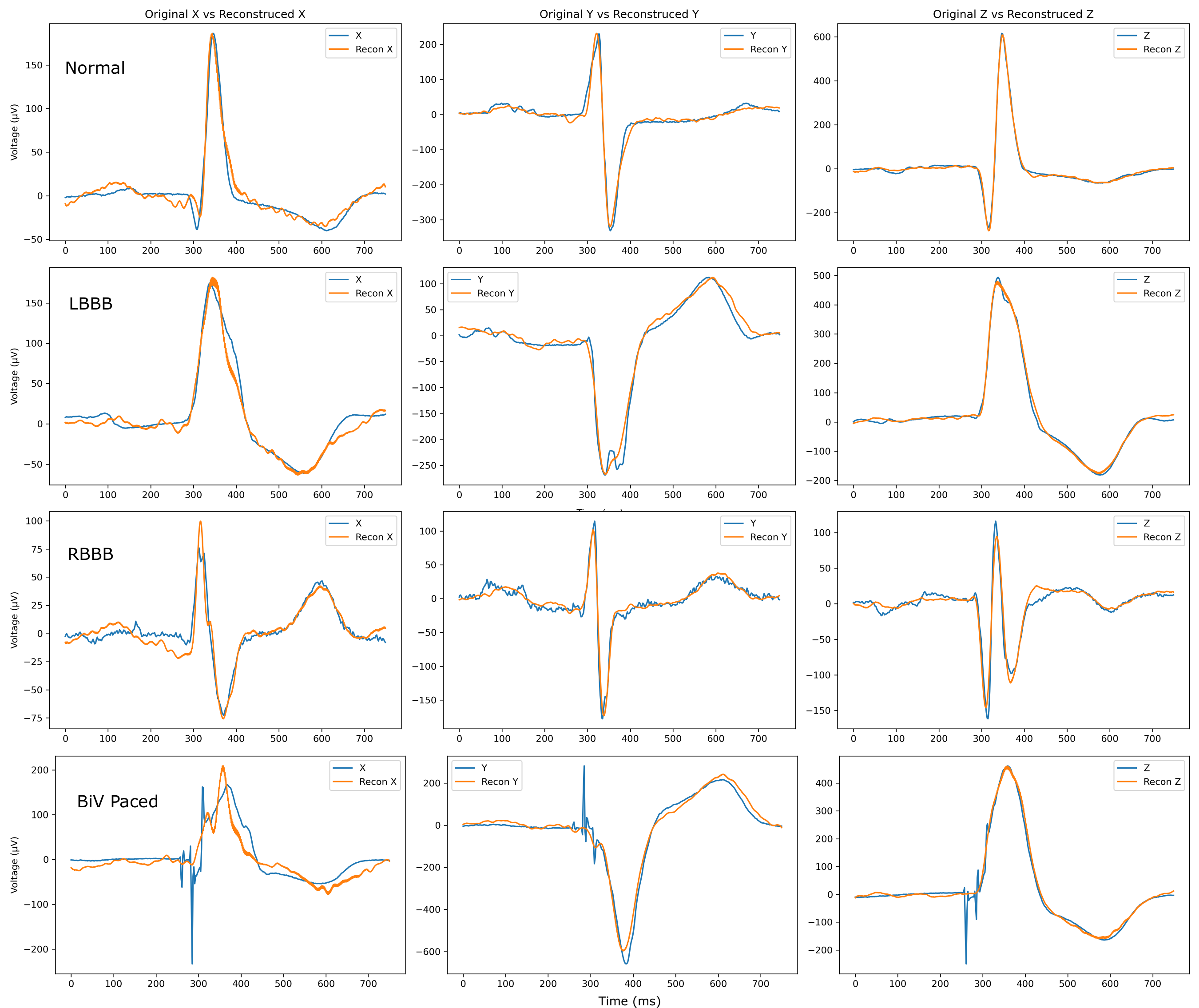}
  
  \caption{Four examples (Normal, LBBB, RBBB and Biventricular Paced) of signal reconstruction (orange) with original signal (blue) overlaid from the A$\beta$-VAE model. The model can reconstruct with a variety of noises and generally smooths out the artifacts.}
  \label{fig:sigrecon}
\end{figure*}

We evaluated seven models—PCA, AE, SAE, VAE, $\beta$-VAE, C$\beta$-VAE, and A$\beta$-VAE—for their ability to reconstruct ECG signals, using Mean Absolute Error (MAE), Mean Squared Error (MSE), and Dynamic Time Warping (DTW) scores shown in Table \ref{tab:MAE_MSE_DTW}. The A$\beta$-VAE achieved the best performance, with an MAE of 15.7$\pm$3.2 µV—in range of baseline signal noise—followed closely by the pure AE, 15.8$\pm$3.1 µV. PCA performed notably worse, with a DTW score over twice that of A$\beta$-VAE. This highlights the efficiency of VAE-based models, particularly A$\beta$-VAE, in preserving ECG signal fidelity.
Reconstructions of X, Y and Z leads from the 30 latent variables showed that the Z lead reconstructed 11.2\% and 12.5\% better than X and Y leads, respectively, possibly due to its unique morphological features related to heart-torso anatomical relationships (Table \ref{tab:MAE_DTW_Signals}). 
Examples of the original signal and the reconstruction from the A$\beta$-VAE model overlaid across varied ECG types are shown in Figure \ref{fig:sigrecon}. 

\subsection{Prediction of QRS Measurements}

\begin{table}[h]
\caption{Prediction in test set of ECG measurements with LGBM using representative beat X, Y, Z-lead ECG encoded variables (n=97,464)}
\centering
\scriptsize
\setlength{\tabcolsep}{3pt} 
\begin{tabular}{lcccccc}
\hline
Model & \multicolumn{2}{c}{QRS Duration (ms)} & \multicolumn{2}{c}{Amplitude$_{\text{QRS-3D}}$ ($\mu V$)} & \multicolumn{2}{c}{VTI$_{\text{QRS-3D}}$ ($\mu Vs$)} \\
\cline{2-3} \cline{4-5} \cline{6-7}
& MAE $\pm$ SD & $R^2$ & MAE $\pm$ SD & $R^2$ & MAE $\pm$ SD & $R^2$ \\
\hline
PCA & \textbf{8.1$\pm$12.9} & \textbf{0.727} & \textbf{108.3$\pm$173.3} & \textbf{0.838} & \textbf{3.16$\pm$5.32} & \textbf{0.923} \\
AE & 8.7$\pm$13.8 & 0.687 & 108.8$\pm$175.3 & 0.835 & 3.68$\pm$5.96 & 0.903 \\
SAE & 8.3$\pm$13.2 & 0.712 & 109.1$\pm$176.6 & 0.835 & 3.42$\pm$5.64 & 0.914 \\
VAE & 8.4$\pm$13.4 & 0.705 & 109.7$\pm$177.6 & 0.833 & 3.46$\pm$5.72 & 0.911 \\
$\beta$-VAE & 8.5$\pm$13.5 & 0.697 & 109.6$\pm$177.4 & 0.833 & 3.54$\pm$5.83 & 0.908 \\
A$\beta$-VAE & 8.3$\pm$13.3 & 0.709 & 107.7$\pm$175.3 & 0.837 & 3.44$\pm$5.68 & 0.912 \\
c$\beta$-VAE & 8.3$\pm$13.3 & 0.708 & 110.2$\pm$177.9 & 0.832 & 3.44$\pm$5.69 & 0.912 \\
\hline
\end{tabular}
\label{tab:ECG_Predictions}
\end{table}

Using the 30 latent variables from each model, we trained LGBM models to predict QRS duration, 3D QRS amplitude, and scalar 3D QRS voltage-time integral (VTI$_{QRS-3D}$). For reference, the average measured QRS duration was 94.8$\pm$29.2 ms, amplitude$_{QRS-3D}$ was 942.8$\pm$434.2 µV and the VTI$_{QRS-3D}$ was 32.12$\pm$17.43 µVs, and these were predicted from the encodings with an error in the range of approximately 10\%.
PCA excelled with an R² of 0.923 for VTI$_{QRS-3D}$, reflecting its ability to extract separable features. Among VAEs, the SAE performed the best with an R² of 0.914, demonstrating that the encodings effectively capture critical signal characteristics (Table \ref{tab:ECG_Predictions}).

\subsection{Prediction of Bundle Branch Blocks}

The right and left bundle branches are the main conduction branches for ventricular activation.
Conduction delay or block in either of these, RBBB or LBBB, lead to increased QRS duration, both with distinctive QRS morphological alterations.
We used the American Heart Association/American College of Cardiology Foundation/Heart Rhythm Society criteria for diagnosis of RBBB and LBBB to develop custom code to identify these conduction abnormalities \cite{Surawicz2009}.  

\begin{table}[h]
\caption{Prediction in test set of RBBB and LBBB with LGBM using representative beat X, Y, Z-lead ECG encoded variables (n=97,464), specificity set at 0.9.}
\centering
\scriptsize
\begin{tabular}{lcccc}
\hline
Model & \multicolumn{2}{c}{RBBB (8.06\% Prevalence)} & \multicolumn{2}{c}{LBBB (3.99\% Prevalence)} \\
\cline{2-3} \cline{4-5}
& AUROC & Sensitivity & AUROC & Sensitivity \\
\hline
PCA & 0.9435 & 0.894 & 0.9637 & 0.939 \\
AE & 0.9390 & 0.881 & 0.9618 & 0.938 \\
SAE & 0.9504 & 0.906 & \textbf{0.9701} & \textbf{0.948} \\
VAE & 0.9507 & 0.904 & 0.9688 & 0.950 \\
$\beta$-VAE & 0.9473 & 0.895 & 0.9689 & 0.949 \\
A$\beta$-VAE & 0.9499 & 0.903 & 0.9686 & 0.947 \\
c$\beta$-VAE & \textbf{0.9516} & \textbf{0.908} & 0.9697 & 0.949 \\
\hline
\end{tabular}
\label{tab:RBBB_LBBB}
\end{table}

LGBM models trained on the 30 encoded variables predicted RBBB and LBBB classification, with VAEs outperforming PCA and AE, Table \ref{tab:RBBB_LBBB}. C$\beta$-VAE achieved the highest AUROC for RBBB (0.9516) and SAE for LBBB (0.9701). Despite imbalanced label distributions, VAEs maintained high sensitivity and specificity, indicating their capability to capture complex QRS morphological variations.

\subsection{Downstream Clinically-relevant Prediction}

The high reconstruction fidelity and predictive power of VAE encodings validate their role as surrogates for raw ECG signals. 
Unlike summary statistics, these encodings preserve subtle morphological details. This enables simpler tree-based (LGBM, Random Forest) or regression models to utilize the encodings effectively and achieve superior predictions compared to models trained on ECG summary statistics.

\begin{table}[h]
\caption{Prediction in test set of reduced LVEF with LGBM using representative beat X, Y, Z-lead ECG encoded variables in test set (n=30,554). Reduced LVEF 14.09\% prevalence, specificity set at 0.9.}
\centering
\scriptsize
\begin{tabular}{lcccc}
\hline
Model & \multicolumn{2}{c}{Reduced LVEF ($\leq$35\%)} & \multicolumn{2}{c}{LVEF, \%} \\
\cline{2-3} \cline{4-5}
& AUROC & Sensitivity & MAE $\pm$ SD & $R^2$ \\
\hline
PCA & 0.799 & 0.616 & \textbf{8.86$\pm$12.15} & \textbf{0.247} \\
AE & 0.810 & 0.656 & 9.05$\pm$12.42 & 0.213 \\
\textbf{SAE} & \textbf{0.820} & \textbf{0.665} & 8.96$\pm$12.28 & 0.231 \\
VAE & 0.819 & 0.676 & 8.95$\pm$12.26 & 0.233 \\
$\beta$-VAE & 0.812 & 0.663 & 9.05$\pm$12.39 & 0.217 \\
A$\beta$-VAE & 0.818 & 0.666 & 8.96$\pm$12.29 & 0.229 \\
c$\beta$-VAE & 0.820 & 0.675 & 8.97$\pm$12.28 & 0.231 \\
ECG statistics& 0.761& 0.554 & 9.55$\pm$12.92 & 0.148 \\
\hline
\end{tabular}
\label{tab:LVEF_Prediction}
\end{table}

To make comparisons, we used LVEF $\leq$35\% as the clinically-relevant downstream prediction. We had 303,265 ECGs on 105,370 patients paired with a unique echocardiogram within 45 days in total for both training and testing sets. Reduced LVEF label was present in 14.4\% of the ECGs.
The SAE encodings yielded the highest AUROC (0.8203) in binary classification, surpassing traditional ECG statistics (AUROC 0.7605) and PCA (AUROC 0.799) shown in Table \ref{tab:LVEF_Prediction}. Combining SAE encodings with traditional ECG features improved the AUROC to 0.901, nearing a state-of-the-art CNN using the original signal, 0.909. With only 10\% training data, the LGBM model maintained an AUROC of 0.870, while the CNN dropped to 0.630, shown in Table \ref{tab:reduced_size}.

\begin{table}[h]
\caption{Performance of different machine learning models in predicting reduced LVEF (LVEF$\leq$35\%) from ECG data (Holdout Test Set: n=15,987), specificity set at 0.9.}
\centering
\scriptsize
\begin{tabular}{lccc}
\hline
Model & Training Sample Size & AUROC & Sensitivity \\
\hline
CNN & 100\% (n=143,644) & \textbf{0.909} & \textbf{0.742} \\
 & ~9.5\% (n=13,568) & 0.630 & 0.177 \\
ResNet & 100\% (n=143,644) & \textbf{0.892} & \textbf{0.672} \\
 & 10\% (n=14,364) & 0.855 & 0.586 \\
 & 1\% (n=1,436) & 0.811 & 0.462 \\
 & 0.1\% (n=143) & 0.705 & 0.281 \\
LGBM & 100\% (n=143,644) & \textbf{0.901} & \textbf{0.702} \\
 & 10\% (n=14,364) & 0.870 & 0.610 \\
 & 1\% (n=1,436) & 0.846 & 0.525 \\
 & 0.1\% (n=143) & 0.761 & 0.361 \\
\hline
\end{tabular}
\label{tab:reduced_size}
\end{table}

These results highlight that VAE encodings allow traditional ML models using substantially less training data and computational resources to perform comparably to DL models trained on full ECG signals. This approach would facilitate the development of ECG diagnostic tools for minority populations, rare health conditions, and invasive modalities where large training datasets are not available. 

\subsection{VAE Performance}

The SAE’s unexpected success at ECG-based downstream prediction tasks suggests that prioritizing reconstruction quality over latent space regularization (via KL divergence) is beneficial. Unlike other VAEs, SAE optimizes solely for reconstruction loss, better capturing ECG variability. However, the pure AE’s inferior performance highlights the value of stochastic sampling in VAEs for generalizability.

\subsection{Discussion}

Our study presents significant advancements in applying autoencoder techniques, particularly our novel VAE variants, to ECG data analysis in clinical diagnostics.
By effectively reducing high-dimensional ECG signals to a compact set of latent variables without requiring extensive datasets, we address a critical challenge in utilizing ECG data in machine learning models.
Our novel VAE variants, particularly the A$\beta$-VAE and SAE, demonstrate superior performance in preserving essential morphological features of the ECG by capturing non-linear relationships within the data, which is crucial for accurate clinical interpretations. Compared to traditional methods like PCA, our models better retain clinically relevant information, as evidenced by the improved prediction of complex ECG features such as bundle branch blocks and reduced LVEF. This enhanced capability leads to improved diagnostic accuracy, which is particularly impactful for conditions like heart failure where early detection can significantly alter patient outcomes.

By accommodating the inherent variability of ECG signals among individuals—through the SAE's focus on reconstruction quality over latent space regularization—our approach allows for more personalized and accurate assessments, aligning with the move towards personalized medicine.
Additionally, enabling robust predictive models in environments with limited datasets contributes to more equitable healthcare delivery by providing advanced diagnostic capabilities across under-represented and diverse populations.
This extends the utility of VAEs in ECG analysis beyond interpretability to practical predictive performance, offering a pathway to integrate these models into existing clinical workflows even with smaller datasets.

\section{Conclusion and Future work}
This study demonstrates that ECG data, despite its inherent complexity and variability, can be effectively reduced using PCA and VAEs for a wide range of downstream prediction tasks. Our approach shows that 60,000 data points in a full 10-sec 12-lead ECG at 500 Hz can be reduced to 30 latent variables with minimal information loss.
While PCA remains a strong contender for basic feature extraction, its limitations become apparent in more complex prediction tasks where access to the full range of ECG signal details is crucial.
In these scenarios, the novel VAE variants, particularly the SAE and C$\beta$-VAE, demonstrate high-fidelity signal reconstructions and accurate downstream predictions.
The SAE encodings stood out, excelling in all prediction tasks while also providing good signal reconstruction.
This finding challenges the conventional wisdom that regularizing the latent space is always beneficial, suggesting instead that a focus on reconstruction quality can yield equally, if not more, valuable results.
In addition, VAEs can be used for synthetic signal data generation. 

The deficiency of our method is that the representative-beat encodings lack the beat-to-beat information as they are created from one representative heartbeat.
The 10-sec data is crucial for cardiac rhythm and arrhythmic detection and additionally captures information on autonomic nervous function and susceptibility to arrhythmogenesis. 
The next frontier for complexity reduction research for ECGs is to encode the full 10-sec signal.

By continuing to refine these encoding techniques, we aim to create more robust diagnostic tools that can be applied to minority populations, rare health conditions and invasive cardiac procedures, ultimately enhancing the accessibility and effectiveness of ECG-based diagnostics. In future work, we will explore clinical applications of these methods in such prediction tasks that currently lack large-scale training datasets.


\section{Acknowledgements}

This work was supported by the Department of Cardiovascular Medicine at The University of Kansas Medical Center, the American Heart Association (AHA) Transformational Project Award (https://doi.org/10.58275/AHA.24TPA1291852.pc.gr.19 6660) to Amit Noheria, and Clinical and Translational Science Award (CTSA) grant from National Center for Advancing Translational Sciences (NCATS) awarded to The University of Kansas for Frontiers: University of Kansas Clinical and Translational Science Institute (UL1TR002366).The content is solely the responsibility of the authors and does not necessarily represent the official views of the AHA, NCATS or National Institutes of Health (NIH).


\setlength{\bibsep}{2.5pt plus 0ex}
\footnotesize
\bibliographystyle{IEEEbibSPMB}
\bibliography{IEEEabrv,IEEESPMB}

\begin{thebibliography}{10}
\providecommand{\url}[1]{#1}
\csname url@samestyle\endcsname
\providecommand{\newblock}{\relax}
\providecommand{\bibinfo}[2]{#2}
\providecommand{\BIBentrySTDinterwordspacing}{\spaceskip=0pt\relax}
\providecommand{\BIBentryALTinterwordstretchfactor}{4}
\providecommand{\BIBentryALTinterwordspacing}{\spaceskip=\fontdimen2\font plus
\BIBentryALTinterwordstretchfactor\fontdimen3\font minus \fontdimen4\font\relax}
\providecommand{\BIBforeignlanguage}[2]{{%
\expandafter\ifx\csname l@#1\endcsname\relax
\typeout{** WARNING: IEEEtran.bst: No hyphenation pattern has been}%
\typeout{** loaded for the language `#1'. Using the pattern for}%
\typeout{** the default language instead.}%
\else
\language=\csname l@#1\endcsname
\fi
#2}}
\providecommand{\BIBdecl}{\relax}
\BIBdecl

\bibitem{berkaya2018survey}
S.~K. Berkaya, A.~K. Uysal, E.~S. Gunal, S.~Ergin, S.~Gunal, and M.~B. Gulmezoglu, ``A survey on ecg analysis,'' \emph{Biomedical Signal Processing and Control}, vol.~43, pp. 216--235, 2018.

\bibitem{kumar2022classification}
A.~Kumar~M and A.~Chakrapani, ``Classification of ecg signal using fft based improved alexnet classifier,'' \emph{PLOS one}, vol.~17, no.~9, p. e0274225, 2022.

\bibitem{DASAN2021102225}
\BIBentryALTinterwordspacing
E.~Dasan and I.~Panneerselvam, ``A novel dimensionality reduction approach for ecg signal via convolutional denoising autoencoder with lstm,'' \emph{Biomedical Signal Processing and Control}, vol.~63, p. 102225, 2021.   (available at: \emph{\url{https://www.sciencedirect.com/science/article/pii/S1746809420303554}}).
\BIBentrySTDinterwordspacing

\bibitem{kors1990reconstruction}
J.~Kors, G.~Van~Herpen, A.~Sittig, and J.~Van~Bemmel, ``Reconstruction of the frank vectorcardiogram from standard electrocardiographic leads: diagnostic comparison of different methods,'' \emph{European heart journal}, vol.~11, no.~12, pp. 1083--1092, 1990.

\bibitem{Murphy2010}
S.~N. Murphy, G.~Weber, M.~Mendis, V.~Gainer, H.~C. Chueh, S.~Churchill, and I.~Kohane, ``Serving the enterprise and beyond with informatics for integrating biology and the bedside ({i2b2}),'' \emph{Journal of the American Medical Informatics Association}, vol.~17, no.~2, pp. 124--130, 2010.

\bibitem{Waitman2011}
L.~R. Waitman, J.~J. Warren, E.~L. Manos, and D.~W. Connolly, ``Expressing observations from electronic medical record flowsheets in an {i2b2}-based clinical data repository to support research and quality improvement,''   \emph{AMIA Annual Symposium Proceedings}, 2011, pp. 1454--1463.

\bibitem{kingma2013auto}
D.~P. Kingma and M.~Welling, ``Auto-encoding variational bayes,'' \emph{arXiv preprint arXiv:1312.6114}, 2013.

\bibitem{kullback1951information}
S.~Kullback and R.~A. Leibler, ``On information and sufficiency,'' \emph{The annals of mathematical statistics}, vol.~22, no.~1, pp. 79--86, 1951.

\bibitem{higgins2017betavae}
\BIBentryALTinterwordspacing
I.~Higgins, L.~Matthey, A.~Pal, C.~Burgess, X.~Glorot, M.~Botvinick, S.~Mohamed, and A.~Lerchner, ``beta-{VAE}: Learning basic visual concepts with a constrained variational framework,''   \emph{International Conference on Learning Representations}, 2017.   (available at: \emph{\url{https://openreview.net/forum?id=Sy2fzU9gl}}).
\BIBentrySTDinterwordspacing

\bibitem{fu2019cyclical}
H.~Fu, C.~Li, X.~Liu, J.~Gao, A.~Celikyilmaz, and L.~Carin, ``Cyclical annealing schedule: A simple approach to mitigating kl vanishing,'' \emph{arXiv preprint arXiv:1903.10145}, 2019.

\bibitem{attia2019screening}
Z.~I. Attia, S.~Kapa, F.~Lopez-Jimenez, P.~M. McKie, D.~J. Ladewig, G.~Satam, P.~A. Pellikka, M.~Enriquez-Sarano, P.~A. Noseworthy, T.~M. Munger \emph{et~al.}, ``Screening for cardiac contractile dysfunction using an artificial intelligence--enabled electrocardiogram,'' \emph{Nature medicine}, vol.~25, no.~1, pp. 70--74, 2019.

\bibitem{attia2019age}
Z.~I. Attia, P.~A. Friedman, P.~A. Noseworthy, F.~Lopez-Jimenez, D.~J. Ladewig, G.~Satam, P.~A. Pellikka, T.~M. Munger, S.~J. Asirvatham, C.~G. Scott \emph{et~al.}, ``Age and sex estimation using artificial intelligence from standard 12-lead ecgs,'' \emph{Circulation: Arrhythmia and Electrophysiology}, vol.~12, no.~9, p. e007284, 2019.

\bibitem{cho2021artificial}
J.~Cho, B.~Lee, J.-M. Kwon, Y.~Lee, H.~Park, B.-H. Oh, K.-H. Jeon, J.~Park, and K.-H. Kim, ``Artificial intelligence algorithm for screening heart failure with reduced ejection fraction using electrocardiography,'' \emph{ASAIO Journal}, vol.~67, no.~3, pp. 314--321, 2021.

\bibitem{10.5555/3294996.3295074}
G.~Ke, Q.~Meng, T.~Finley, T.~Wang, W.~Chen, W.~Ma, Q.~Ye, and T.-Y. Liu, ``Lightgbm: A highly efficient gradient boosting decision tree,''   \emph{Proceedings of the 31st International Conference on Neural Information Processing Systems}, ser. NIPS'17. Red Hook, NY, USA: Curran Associates Inc., 2017, p. 3149–3157.

\bibitem{Surawicz2009}
B.~Surawicz, R.~Childers, B.~J. Deal, L.~S. Gettes, J.~J. Bailey, A.~Gorgels, E.~W. Hancock, M.~Josephson, P.~Kligfield, J.~A. Kors, P.~Macfarlane, J.~W. Mason, D.~M. Mirvis, P.~Okin, O.~Pahlm, P.~M. Rautaharju, G.~van Herpen, G.~S. Wagner, H.~Wellens, {American Heart Association Electrocardiography and Arrhythmias Committee, Council on Clinical Cardiology}, {American College of Cardiology Foundation}, and {Heart Rhythm Society}, ``{AHA/ACCF/HRS recommendations for the standardization and interpretation of the electrocardiogram: part {III}: intraventricular conduction disturbances: a scientific statement from the {American Heart Association} {Electrocardiography and Arrhythmias Committee}, {Council on Clinical Cardiology}; the {American College of Cardiology Foundation}; and the {Heart Rhythm Society}},'' \emph{Journal of the American College of Cardiology}, vol.~53, no.~11, pp. 976--981, Mar 2009, endorsed by the International Society for Computerized Electrocardiology.

\end{thebibliography}

\end{document}